\documentclass[runningheads]{llncs}
\usepackage{graphicx}
\usepackage{amsmath}
\usepackage{amssymb}
\usepackage{multirow}
\begin{document}
\title{Copy-Enhanced Heterogeneous Information Learning for Dialogue State Tracking}

\author{Qingbin Liu\inst{1,2} \and
Shizhu He\inst{1} \and
Kang Liu\inst{1}\and
Shengping Liu\inst{3}\and
Jun Zhao\inst{1,2}}

\authorrunning{Qingbin Liu et al.}

\institute{National Laboratory of Pattern Recognition, Institute of Automation, Chinese Academy of Sciences \and
University of Chinese Academy of Sciences \and
Beijing Unisound Information Technology Co., Ltd \\
\email{\{qingbin.liu,shizhu.he,kliu,jzhao\}@nlpr.ia.ac.cn},\\liushengping@unisound.com}
\titlerunning {Copy-Enhanced Heterogeneous Information Learning for DST}
\maketitle
\begin{abstract}
Dialogue state tracking (DST) is an essential component in task-oriented dialogue systems, which estimates user goals at every dialogue turn. However, most previous approaches usually suffer from the following problems. Many discriminative models, especially end-to-end (E2E) models, are difficult to extract unknown values that are not in the candidate ontology; previous generative models, which can extract unknown values from utterances, degrade the performance due to ignoring the semantic information of pre-defined ontology. Besides, previous generative models usually need a hand-crafted list to normalize the generated values. How to integrate the semantic information of pre-defined ontology and dialogue text (heterogeneous texts) to generate unknown values and improve performance becomes a severe challenge. In this paper, we propose a Copy-Enhanced Heterogeneous Information Learning model with multiple encoder-decoder for DST (CEDST), which can effectively generate all possible values including unknown values by copying values from heterogeneous texts. Meanwhile, CEDST can effectively decompose the large state space into several small state spaces through multi-encoder, and employ multi-decoder to make full use of the reduced spaces to generate values. Multi-encoder-decoder architecture can significantly improve performance. Experiments show that CEDST can achieve state-of-the-art results on two datasets and our constructed datasets with many unknown values.

\end{abstract}
%
\section{Introduction}
\noindent Task-oriented dialogue systems help users to achieve specific goals such as finding restaurants \cite{wen-EtAl:2017:EACLlong} and movie information retrieval \cite{dhingra-EtAl:2017:Long1}. DST is a core component in task-oriented dialogue systems. DST estimates user goals through the dialogue context. Dialogue systems use these user goals to decide the next system actions, which are used to generate natural language responses.

\begin{figure}
\centering\includegraphics[width=2.8in]{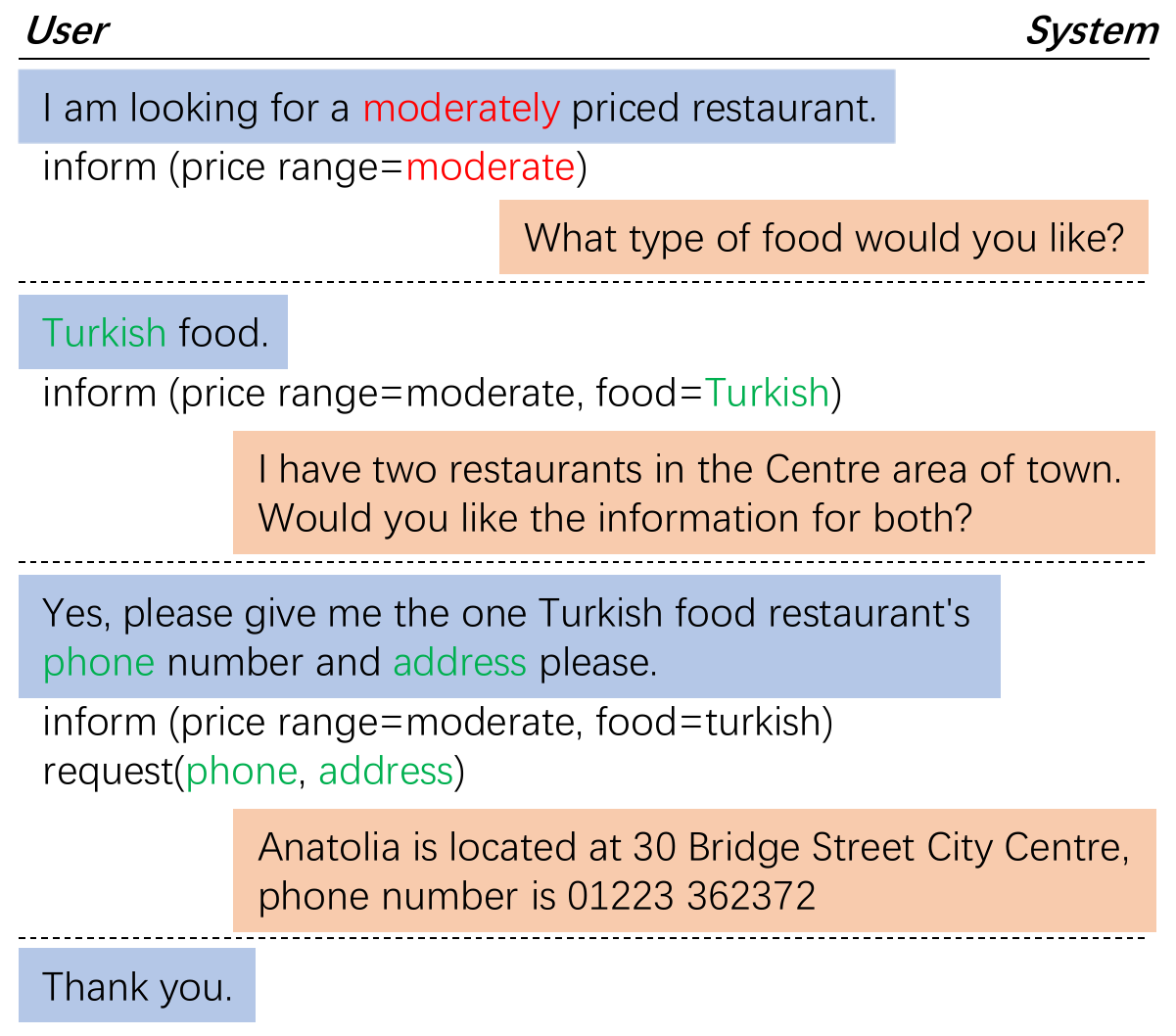}
\caption{An example of the DST in finding restaurant task. Given each system response (orange) and the following user utterance (blue), DST estimates joint and request goals. The red-color word shows rephrasing of values. The green-color word can be copied from the dialogue context directly as unknown values.} \label{fig1}
\end{figure}

Typically, in DST tasks, all user goals are in a fixed domain ontology and represented by slot-value pairs. Consider the task of finding restaurants in Figure \ref{fig1} as an example. During each turn, the user informs specific constraints (e.g. inform(price range = moderate)) or requests some information (e.g. request(address)). Constraints informed in one turn is called turn goals. Joint goals are the constraints informed in current and previous turns. These values vary lexically and morphologically (e.g. moderate: [moderately, mid-priced]) as shown in the red-color word in Figure 1. The rephrasing of values needs to be converted into a standard form in the fixed ontology. Therefore, traditional DST is to extract possible the standard slot-value pairs given the conversation context.

In the research community, previous models with a fixed ontology are not scalable in many domains, because 1) the entire dataset is usually not exposed to us and we can only get a part of the values from the training set, and 2) such a fixed ontology is not scalable when the value set is dynamic (e.g. new movie added, new restaurant opened, etc.) and unbounded (e.g. date, volume etc.) \cite{DBLP:journals/corr/abs-1712-10224}. The scalability of these models in real-world dialogues remains to be evaluated.

Dealing with the values that don't appear in the ontology is a critical but rarely mentioned capability of DST. The generative models with pointer or copy-mechanism can generate the unknown value by selecting words in the dialogue as values \cite{sutskever2014sequence}. However, the performance of common generative models is low due to weakly considering the semantic information of the ontology. For example, \cite{P18-1134} pointers out text spans from the dialogue as unknown values without referring to the ontology, which could not learn the non-pointer values. Besides, they used a hand-crafted list to exhaustively enumerate all rephrasing for each value. The non-pointer and non-enumerated values degrade performance.

In this paper, we propose a Copy-Enhanced Heterogeneous Information Learning model (CEDST), which can generate all the possible value including the known and unknown value from the heterogeneous texts (dialogue context and ontology). Based on the findings that most values (e.g. 82\% in the WoZ2 dataset) can be copied directly from the conversation as shown in the green-color word in Figure \ref{fig1}. Our proposed model is augmented with the copy mechanism \cite{gu2016incorporating} to only copy values with the normalized form. Moreover, CEDST maps the heterogeneous known values and dialogues into multiple vector spaces by multi-encoder and uses multi-decoder to extract dialogues states from these reduced spaces. The multi-encoder-decoder architecture extracts the slot-related and slot-shared features. This architecture can effectively generate a reduced state space for each slot and learn contextual information to generate the unknown values. Since there are no public DST datasets with the unknown value, we construct datasets containing the unknown values in different proportions. CEDST is able to achieve state-of-the-art results on two public and constructed datasets.

In summary, this paper makes the following contributions:
\begin{itemize}
\item We apply copy mechanism in DST tasks. Through this mechanism, CEDST effectively copies unknown values from the conversation for each slot.
\item We propose the multi-encoder-decoder architecture in CEDST to decompose the heterogeneous texts into multiple small spaces. The value for each slot is effectively selected from the corresponding space.
\item Many experiments are constructed to evaluate the generative capability of models. And, we construct DST datasets with the unknown value in the development and test sets. CEDST is able to achieve state-of-the-art performance on these datasets.
\end{itemize}
\section{Related Work}

Current dialogue state trackers usually employ the E2E models. Typically, E2E models estimate the turn-level user goal and request given a system response and following user utterance at every turn, and then estimates the joint goal by considering all previous dialogue turns.

\subsection{Discriminative Models in DST}

Typically, Discriminative models involve multi-class or multiple binary classifications to model the posterior probability of every candidate value. \cite{henderson2014word} employs a strategy called delexicalisation, which uses hand-crafted semantic dictionaries to replace slots and values mentioned in dialogues with generic labels. Therefore, the corpus is converted to this form such as ('want a \textit{tagged-value} restaurant.'). Many $n$-gram features are collected from the converted corpus. These fratures are used by the model to make a binary decision for every slot-value pair \cite{henderson2014word}. The neural belief tracker \cite{mrkvsic-EtAl:2017:Long} automatically learns effective features from word embedding without semantic dictionaries in the delexicalisation strategy. Memory network is also applied to automatically learn effective features \cite{perez-liu:2017:EACLlong}. The global-locally self-attentive model proposed by \cite{P18-1135} extracts local features in each slot and global features between slots. These local and global features enable the model to focus more on rare slot-value pairs.

As far as we know, \cite{DBLP:journals/corr/abs-1712-10224} is the only work using the discriminative model to handle the dynamic and unbounded value set. \cite{DBLP:journals/corr/abs-1712-10224} represents the dialogue states by candidate sets derived from the dialogue and knowledge, then scores values in the candidate set with binary classifications. Although the sophisticated generation strategy of the candidate set allows the model to extract unknown values, \cite{DBLP:journals/corr/abs-1712-10224} needs a separate SLU module and may propagate errors.

\subsection{Generative Models in DST}

Generative models with the pointer or copy mechanism can generate unknown values. \cite{P18-1134} uses the pointer network to select a continuous text span on the system response and following user utterance as final dialogue states. However, it cannot deal with discontinuous and unclear values and ignore the semantic information in the ontology. In addition, \cite{P18-1134} needs an additional post-normalization to deal with lexical diversity.
\section{Copy-Augmented Heterogeneous Information Learning}
In this paper, DST is regarded as a E2E generative problem with heterogeneous texts. Given a user utterance ${U} = (u_1, u_2, ..., u_n)$ and a system response $S = (r_1, r_2, ..., r_l)$, the model generates possible values from $U$, $S$ or the pre-defined ontology. In this section, we will firstly introduce a general encoder-decoder framework with copy mechanism \cite{gu2016incorporating}, and then introduce CEDST.
\subsection{Background: Encoder-Decoder with Copy Mechanism}
The encoder-decoder framework was firstly introduced in sequence-to-sequence learning \cite{sutskever2014sequence} and then quickly extended to many NLP tasks. In this framework, an encoder, usually recurrent neural networks such as long-short term memory (LSTM) \cite{hochreiter1997long}, transforms the input sequence $X = (x_1, x_2, ..., x_m)$ to a hidden representation $H = (h_1, h_2, ..., h_m)$ as below:
\begin{equation}
{h_t}={LSTM}({h}_{t-1}, e(x_t))
\end{equation}
where $e(x_t)$ is the word embedding of $x_t$.

The decoder uses the generated word $y_{t-1}$ decoded in the last step and a vector $a_{t-1}$ calculated by attention mechanism \cite{DBLP:journals/corr/BahdanauCB14} to update its states $s_t$:
\begin{equation}
{s_t}={LSTM}({s}_{t-1}, [a_{t-1}; e(y_{t-1})])
\end{equation}
The target word is sampled from the following probability distribution:
\begin{equation}
\begin{split}
y_t \sim o_t =P(y_t|y_{<t}, s_t)
\end{split}
\end{equation}

Copy mechanism allows the encoder-decoder model to copy words directly from the input sequence. The generation probability is as below:
\begin{equation}
\begin{split}
P(y_t|s_t, H)=P(y_t, {\rm \textbf{g}}|s_t, H) + P(y_t, {\rm \textbf{c}}|s_t, H)
\end{split}
\end{equation}
where \textbf{g} represents the generation mode, and \textbf{c} the copy mode.
\subsection{Overview}
\begin{figure}
\centering\includegraphics[width=2.0in]{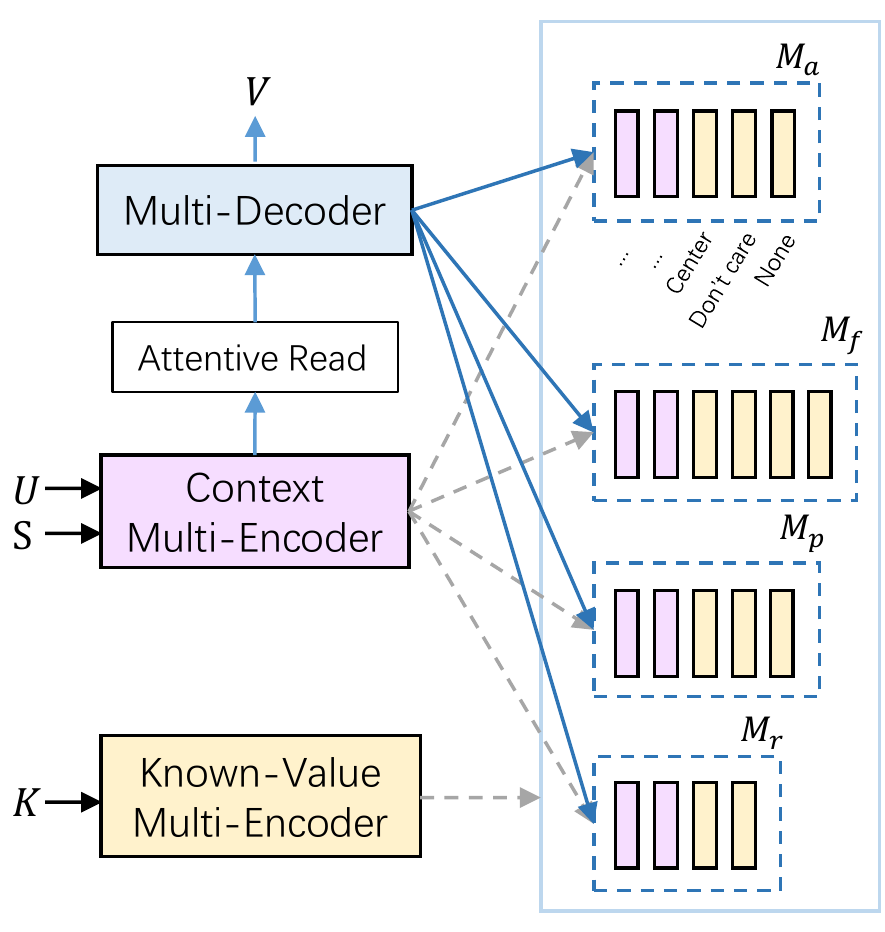}
\caption{The overview of CEDST. The context multi-encoder includes the user utterance ($U$) multi-encoder and the system response ($S$) multi-encoder. $K$ is known values in the training set. $V$ is the generated value for every slot. Multiple memories ($M_a$, $M_f$, $M_p$, $M_r$) is generated according to the corresponding slot.} \label{fig2}
\end{figure}

The overview of CEDST is presented in Figure \ref{fig2}. The context multi-encoder takes the user utterance ($U$) and system response ($S$) as inputs during every turn and generates specific hidden for each slot (purple strips in Figure \ref{fig2}). The hidden is used for the language understanding and copy mechanism. The known-value multi-encoder encodes known values ($K$) in the ontology into the same vector space with the hidden of $U$ and $S$ for each slot. The multi-encoder bridges the representation gap between the words in the dialogue and known values that may consist of multiple words. Because the values of each slot are different, we accordingly generate specific space for each slot. These spaces are consisted of the short-term memories termed as $M_a$, $M_f$, $M_p$, $M_r$ in Figure 2. Multi-Decoder attentively read the hidden and selectively generate the known value or the word step by step for every slot with copy mechanism.

\subsection{Multi-Encoder}
Multi-encoder in CEDST is designed in a private and shared architecture following the multi-encoder in \cite{P18-1135}, which is used for extract features for a discriminator. The private bi-directional LSTM (BiLSTM) focuses on extract the slot-related feature. For example, the words `Chinese restaurant' and `seafood' is likely to be related to the `food' slot. The shared BiLSTM focuses on extract the shared feature between slots such as the affirmative and negative words:
\begin{equation}
{H^s}={BiLSTM^s}(e(X))
\end{equation}
\begin{equation}
{H^g}={BiLSTM^g}(e(X))
\end{equation}
where $H^s, H^g \in \mathbb{R}^{(n \times d_{rnn})}$.
The proportion of the private and shared features is learned by a gate with a trainable parameter.
\begin{equation}
{H}=\beta^sH^s + (1-\beta^s)H^g
\end{equation}
The private self attention learns the related context for each slot.
\begin{equation}
{z^s_i}=W^sH_i + b^s
\end{equation}
\begin{equation}
{p^s}=softmax(z^s)
\end{equation}
\begin{equation}
{c^s}=\sum_ip^s_iH_i
\end{equation}
where $z^s_i, p^s, c^s \in \mathbb{R}, \mathbb{R}^n, \mathbb{R}^{1 \times d_{rnn}}$.
The shared self attention $c^g$ is computed as the above private self attention with the parameters between slots.
The final representation is the integration of private and shared self attention:
\begin{equation}
{c}=\beta^sc^s + (1-\beta^s)c^g
\end{equation}
Finally, multi-encoder transforms each input into a hidden $H$ and a context $c$.

Multi-encoders encode the known values, system response, and user utterance. The hidden makes up the memories (e.g. $M_a$) in CEDST for each slot. Special values such as `don't care' and `none', which indicates that the user doesn't care or has no intent to some slots, are also encoded to a hidden. By doing so, special cases in DST can be processed in a unified framework. The system response is replaced with system actions as previous works. CEDST models the interaction between system actions ($c^a$) and user utterance ($c^u$) as below:
\begin{equation}
{p^{a}}=softmax(c^a(c^u)^T)
\end{equation}
\begin{equation}
c^{n}={\sum}_{i}^lc^a_ip^{a}_i
\end{equation}
where $c^a, c^u, p^a \in \mathbb{R}^{l \times {d_{rnn}}}, \mathbb{R}^{1 \times {d_{rnn}}}, \mathbb{R}^l$.

The slot-specific user utterance, system response, and known values make up the complete state space, hence the large state space of heterogeneous texts is decomposed into multiple spaces. These spaces are closely related to the corresponding slot. For the special values are also in the spaces, every training sample can find the target in these spaces, and this could optimize the inefficient training problem of the data sparseness.

\subsection{Multi-Decoder and Copy Mechanism}
\begin{figure}
\centering\includegraphics[width=3.0in]{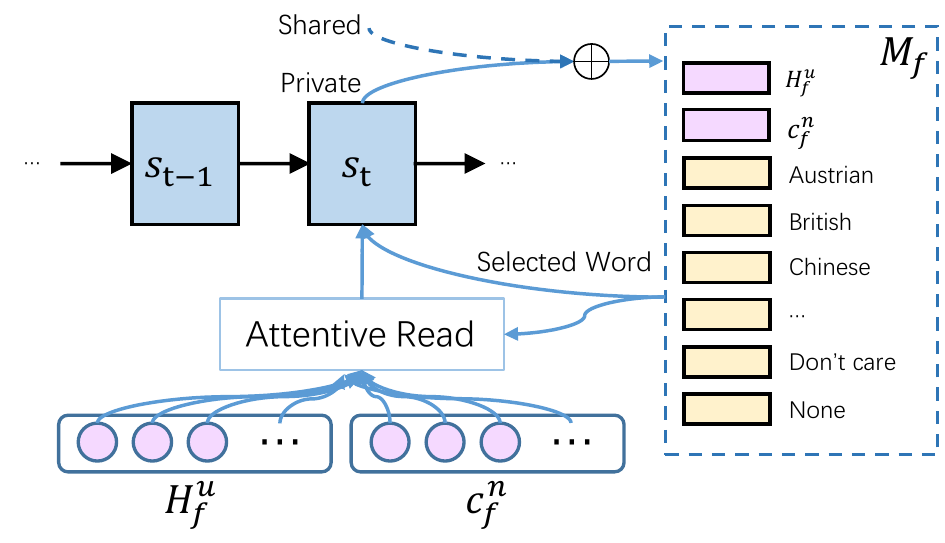}
\caption{The architecture of the multi-decoder, which is generating the value for the `food' slot. $H_f^u$ and $c_f^a$ is the hidden of the user utterance and system response corresponding to this slot. $M_f$ is the short-term memory belong to the `food' slot.} \label{fig3}
\end{figure}

Multi-decoder is designed to effectively utilize these state spaces to generate values for every slot. As shown in Figure \ref{fig3}, multi-decoder employs a private decoder for each slot and a shared decoder between slots to decoder comprehensive information from the inputs. It updates its hidden state ($s_t$) with the last predicted word ($y_{t-1}$) and attention vector ($a_{t-1}$) as below:
\begin{equation}
{s_t^s}={LSTM^s}(s_{t-1}^s, [e(y_{t-1}); a_{t-1}^s])
\end{equation}
\begin{equation}
{s_t^g}={LSTM^g}(s_{t-1}^g, [e(y_{t-1}); a_{t-1}^g])
\end{equation}
\begin{equation}
{s_t}=\gamma^ss_t^s + (1-\gamma^s)s_t^g
\end{equation}
where $\gamma^s$ is a trainable parameter. $a_{t-1}$ is computed as below:
\begin{equation}
{p^{att}}=softmax([H^u; c^{a}](e(y_{t-1}))^T)
\end{equation}
\begin{equation}
a_{t-1}={\sum}_ip^{att}_i[H^u; c^{a}]_i
\end{equation}
where $[H^u; c^{a}]$ is the concatenation of $H^u$ and $c^{a}$.

Some slots only have one target value in a training sample (named as the single-value slot in our paper). For example, the user only inform one value or `none' for the `food' slot in the finding restaurant task. The multi-decoder selects a word or a value for the single-value slot with the following distribution:
\begin{equation}
{p_{t}}=M*(s_t)^T
\end{equation}
\begin{equation}
\begin{split}
P(y_t|s_t, M) = [P(y_t, {\rm \textbf{c}}| s_t, M); P(y_t, {\rm \textbf{g}}| s_t, M)] =softmax(p_t)
\end{split}
\end{equation}
where $M$ is the short-term memory such as $M_a$ or $M_f$. $P(y_t, {\rm \textbf{c}}| s_t, M)$ stands for the probability distribution on the dialogue. $P(y_t, {\rm \textbf{g}}| s_t, M)$ represents the distribution on the ontology. If the model selects the candidate value at the first step, it stops the subsequent decoding and generates the value through $P(y_t, {\rm \textbf{g}}| s_t, M)$. If it copies a word in the utterance $U$ at the first step, it continues decoding until generating the entire value through $P(y_t, {\rm \textbf{c}}| s_t, M)$. Therefore, the generate and copy mode are determined at the first step.

The slot such as the `request' slot has multiple values (the multi-value slot). The common decoder is different to handle the multi-value slot, due to the insufficient coverage problem. This problem may make the model not fully generating all values for the slot. Therefore, our multi-decoder generates multiple words or values only in the first step to alleviate the problem as below:
\begin{equation}
{p}=M*(s_1)^T
\end{equation}
\begin{equation}
\begin{split}
P(y|s_1, M) = [P(y, {\rm \textbf{c}}| s_1, M); P(y, {\rm \textbf{g}}| s_1, M)] =sigmoid(p)
\end{split}
\end{equation}
The words and values will be selected in the first step as long as its probability is greater than a threshold (0.5 in our model). The copy mechanism in the multi-value slot can copy all possible words at one time. The predicted values can be easily obtained through some simple segmentation rules.

The target of copied values is the index of corresponding words in the dialogue, otherwise the index of values in the ontology. We utilize the cross-entropy to optimize the probability.

\section{Experiments}
\subsection{Dataset}
\subsubsection{Public Datasets}
The Wizard of Oz 2 (WoZ2) dataset is used for finding restaurants around Cambridge \cite{wen-EtAl:2017:EACLlong,mrkvsic-EtAl:2017:Long}. WoZ2 contains three single-value informable slots ('area', 'food', 'price range') and a multi-value request slot. API calls in Task 5 of the bAbI dataset is regarded as the state \cite{DBLP:journals/corr/BordesW16}. The out-of-vocabulary (OOV) test set in bAbI only contains unknown values.

\subsubsection{Constructed Datasets with Unknown Values} The standard DST dataset WoZ2 only can evaluate the capability of models to process known values. Therefore, we manually construct datasets based on WoZ2. We randomly select known values in WoZ2 and mask them and their annotation in the train set according to the ratio of 20\%, 40\%, 60\%. Selected values are seamed as unknown values.
\subsection{Implementation Details}
GloVe \cite{Pennington2014Glove} and character n-gram embedding \cite{hashimoto2016joint} is the fixed word embedding. The loss is optimized by ADAM \cite{kingma2014adam} with a fixed learning rate 0.001. The parameters of our model is initialized by the Xavier initializer. We apply dropout \cite{srivastava2014dropout} with 0.8 keep rate to word embedding and outputs of multi-encoder.
\subsection{Comparison on Public Datasets}
\begin{table}
\centering
\caption{Results on WoZ2. Baselines are delexicalisation-based model \cite{henderson2014word}, delexicalisation with semantic dictionaries \cite{wen-EtAl:2017:EACLlong},  NBT \cite{mrkvsic-EtAl:2017:Long}, GLAD \cite{P18-1135} and PtrNet\cite{P18-1134}.}\label{tab1}
\begin{tabular}{|l|c|c|}
\hline
Models & Joint Goal &Turn Request\\
\hline
Delex.        &70.8\%&87.1\%\\
Delex. + Dictionaries  &83.7\%&87.6\%\\
NBT - DNN   &84.4\%&91.2\%\\
NBT - CNN   &84.2\%&91.6\%\\
GLAD                                &88.1 $\pm$ 0.4\%& 97.1 $\pm$ 0.2\%\\
\hline
PtrNet                              &87.5\%     &- \\
CEDST                               & \textbf{89.6\%}& 97.0\%\\
\hline
\end{tabular}
\end{table}

\begin{table}[]
\centering
\caption{Results on bAbI. Comparison with the generative model PtrNet.}\label{tab2}
\begin{tabular}{|l|c|c|c|c|c|}
\hline
\multirow{2}{*}{Model}  & \multicolumn{1}{c|}{\multirow{2}{*}{Slot}} & \multicolumn{2}{l|}{Regular Test} & \multicolumn{2}{l|}{OOV Test} \\ \cline{3-6}
                        & \multicolumn{1}{c|}{}                      & p=0    & p=0.1                    & p=0           & p=0.1         \\ \hline
\multirow{2}{*}{PtrNet} & Food                                       & 100\%    & 100\%                      & 86.2\%          & 100\%           \\ \cline{2-6}
                        & Location                                   & 100\%    & 100\%                      & 74.7\%          & 99.6\%          \\ \hline
\multirow{2}{*}{CEDST}  & Food                                       & 100\%    & -                          & 100\%           & -             \\ \cline{2-6}
                        & Location                                   & 100\%    & -                          & 100\%           & -             \\ \hline
\end{tabular}
\end{table}

\begin{table*}[t]
\centering
\caption{The performance of models on constructed datasets. The ratios of unknown values (UNK) in the three datasets are 20\%, 40\%, 60\%.}\label{tab3}
\begin{tabular}{|l|cc|cc|cc|}
\hline
       & \multicolumn{2}{c|}{20\% UNK}     & \multicolumn{2}{c|}{40\% UNK}     & \multicolumn{2}{c|}{60\% UNK}  \\
Models & Joint Goal      & Turn Request    & Joint Goal      & Turn Request    & Joint Goal      & Turn Request \\ \hline
GLAD   & 51.9\%          & 91.0\%          & 29.2\%          & 75.3\%          & 11.0\%          & 73.9\%       \\ \hline
CEDST  & \textbf{53.6\%} & \textbf{91.2\%} & \textbf{32.4\%} & \textbf{75.6\%} & \textbf{11.2\%} & 73.9\%       \\ \hline
\end{tabular}
\end{table*}
Table 1 shows the performance of models on WoZ2. Delex. is the delexicalisation-based model \cite{henderson2014word}. Delex. + Dictionaries is the delexicalisation-based model with semantic dictionaries to hand the rephrasing \cite{wen-EtAl:2017:EACLlong}. \cite{mrkvsic-EtAl:2017:Long} provides two neural belief trackers that learn features from the word embedding. GLAD utilizes the global-locally encoder and self attention to extract comprehensive features \cite{P18-1135}. The metrics is the joint goal accuracy and turn request. The joint goal is obtained by a simple rule to integrate the current and previous turn goals in CEDST. CEDST uses the value of the previous turn goal if the current turn does not get a new value, or replace the old value with a new one if obtained in current turn.

On WoZ2, CEDST achieves the state-of-the-art result. The joint goal accuracy is 89.6\%. The turn request accuracy is 97.0\%. It reveals that CEDST is good at generating known values. This also proved that the generation and utilization of the reduced multiple state spaces by the multi-encoder-decoder in CEDST is very effective. This architecture can significantly improve performance.

On bAbI, CEDST outperforms PtrNet by 13.8 and 25.3 on the `food' and `location' slots. It proves that CEDST is very effective at generating unknown values. Even PtrNet adopted Targeted Feature Dropout to enhance the generative ability, CEDST still gets the same or better performance.

\subsection{Comparison on Constructed Datasets}
For GLAD \cite{P18-1135} achieves high performance on WoZ2, we compare GLAD with CEDST on constructed datasets. As shown in Table \ref{tab2}, CEDST also achieves state-of-the-art performance. We can see that the two models achieve much lower performance on constructed datasets than on WoZ2 even with only 20\% unknown values. It shows that dealing with unknown values is an important capability of DST models. CEDST increases by 1.7\%, 3.2\% and 0.2\% respectively. In the dataset with 60\% unknown values, both models only get very little labeling information from the training set, which makes the performance very low.
\begin{figure}
\centering\includegraphics[width=2.5in]{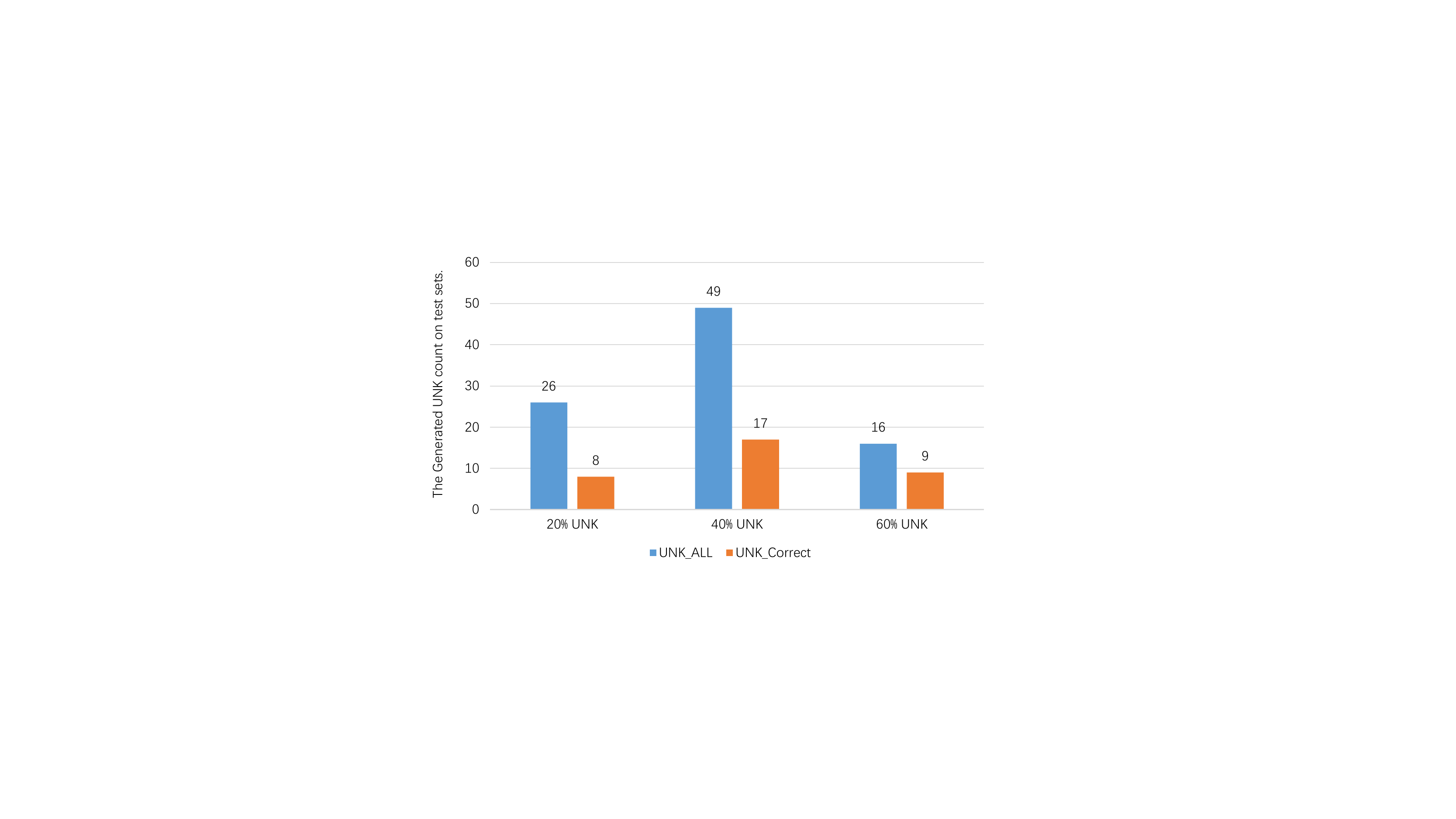}
\caption{The count of generated unknown values. UNK\_ALL is the total number of generated unknown values. UNK\_Correct is the count of correct unknown values.} \label{fig4}
\end{figure}

In order to clearly analyze the improvement of CEDST, we count generated and correct unknown values on the test set as shown in Figure \ref{fig4}. CEDST generates most correct unknown values on the 40\% dataset. Because known values in the 20\% dataset are much more than unknown values, CEDST focuses more on the target word itself and less on context words. The works in \cite{hinton2012improving} and \cite{xu2014targeted} have proved that context words are essential for generating unknown values. On the 60\% dataset, CEDST cannot get sufficient information from the train set and generate less unknown values. The proportion of correct to generated unknown values is 30.8\%, 34.7\% and 56.3\%, which proves the good scalability of CEDST.
\subsection{Ablation Study}

\begin{table*}[t]
\centering
\caption{Ablation experiments on WoZ2 and the 40\% dataset. For `- multi-encoder', we use a single BiLSTM to replace the multi-encoder. For `-multi-decoder', we only use the shared decoder. For `- self att.', we use the last hidden to replace self-attention.  }\label{tab6}
\begin{tabular}{|l|cc|cc|}
\hline
               & \multicolumn{2}{c|}{40\% UNK} & \multicolumn{2}{c|}{The WoZ2 dataset} \\
Models         & Joint Goal   & Turn Request   & Joint Goal       & Turn Request      \\ \hline
CEDST          & 32.4\%       & 75.6\%         & 89.6\%           & 97.0\%            \\
- multi-encoder & 30.6\%       & 75.3\%         & 85.8\%           & 97.0\%            \\
- multi-decoder & 31.2\%       & 75.3\%         & 87.9\%           & 97.1\%            \\
- copy         & 29.4\%       & 75.1\%         & 87.1\%           & 96.2\%            \\
- self att.    & 23.0\%       & 75.3\%         & 70.0\%           & 95.5\%            \\
- shared LSTM   & 31.8\%       & 75.7\%         & 88.6\%           & 97.3\%            \\ \hline
\end{tabular}
\end{table*}

\subsubsection{Multi-Encoder Converts Inputs into Effective Hidden}
We use a single BiLSTM to replace multi-encoder. The BiLSTM only extract shared features and degrades the performance. Private features are very important. For example, the `part' word is only related to the `area' slot. Multi-encoder extract private and shared features, which learn very effective representations.
\subsubsection{Multi-decoder Derives More Useful Information}
The private and shared decoders can derive more information from state spaces. Since the target value and the state space is different, the private decoder that dedicates on decoding specific information for each slot is very useful.
\subsubsection{Copy Mechanism Generates Unknown Values Effectively}
We remove copy mechanism and utterance hidden in $M$. It disables the generation of unknown values and significantly reduces the performance on the 40\% dataset. Because we also remove the utterance hidden, this model degrades the performance on WoZ2. This shows that these hidden can promote dialogue understanding.
\subsubsection{Self Attention}
Self attention in multi-encoder is used to obtain a context hidden. We use the last hidden of the encoder to replace the context. It significantly degrades performance. It shows that self attention can get a context effectively.
\subsubsection{Shared LSTM}
The shared LSTM is removed in multi-encoder-decoder. It only reduces 0.6\% joint goal accuracy on the 40\% dataset and 1.0\% on WoZ2. It even achieves a little higher on the request accuracy. It proved that, in these datasets, global features are less important than private features.

\section{Conclusions}
We propose the copy-augmented heterogeneous information learning model for dealing with unknown values in DST. The copy mechanism in CEDST can copy words in the dialogue context as unknown values. Meanwhile, the multi-encoder-decoder in CEDST can effectively decompose heterogeneous texts into multiple small spaces corresponding to the slots. CEDST can effectively generate values from reduced spaces. CEDST achieves state-of-the-art performance both on WoZ2, bAbI, and our constructed datasets.
\section*{Acknowledge}
This work is supported by the National Natural Science Foundation of China (No.61533018), the Natural Key R\&D Program of China (No.2018YFC0830101), the National Natural Science Foundation of China (No.61702512, No.61806201) and the independent research project of National Laboratory of Pattern Recognition. This work was also supported by Alibaba Group through Alibaba Innovative Research (AIR) Program, CCF-DiDi BigData Joint Lab and CCF-Tencent Open Research Fund.

\bibliographystyle{splncs04}
\bibliography{mybib}

\end{document}